\documentclass[a4paper,twoside]{article}

\usepackage{epsfig}
\usepackage{subcaption}
\usepackage{calc}
\usepackage{amssymb}
\usepackage{amstext}
\usepackage{amsmath}
\usepackage{amsthm}
\usepackage{multicol}
\usepackage{pslatex}
\usepackage{apalike}
\usepackage{acronym}
\usepackage[T1]{fontenc}
\usepackage[utf8]{inputenc}
\usepackage{multirow}
\usepackage{SCITEPRESS}   % Please add other packages that you may need BEFORE the SCITEPRESS.sty package.

\makeglossary

\newacro{ASAG}{Automatic Short Answer Grading}
\newacro{STS}{Semantic Text Similarity}
\newacro{ELMo}{Embeddings from Language Models}
\newacro{BERT}{Bidirectional Encoder Representations from Transformers}
\newacro{GPT}{Generative Pretraining}
\newacro{SVM}{Support Vector Machine}
\newacro{LSA}{Latent Semantic Analysis}
\newacro{BLEU}{Bi-lingual Evaluation Understudy}
\newacro{NLP}{Natural Language Processing}
\newacro{LM}{Language Modeling}
\newacro{LSTM}{Long Short-Term Memory}
\newacro{MLM}{Masked Language Model}
\newacro{SOWE}{Sum of Word Embeddings}
\newacro{RMSE}{Root Mean Square Error}
\newacro{BOW}{Bag-of-Words}
\newacro{tf-idf}{term frequency–inverse document frequency}
\newacro{NLP}{Natural Language Processing}

\begin{document}

\title{Comparative Evaluation of Pretrained Transfer Learning Models on Automatic Short Answer Grading}
%Authors' Instructions: Preparation of Camera-Ready Contributions to SCITEPRESS Proceedings
\author{\authorname{Sasi Kiran Gaddipati, Deebul Nair and Paul G. Plöger}
\affiliation{Hochschule Bonn-Rhein-Sieg, Germany}
\email{sasi-kiran.gaddipati@smail.inf.h-brs.de, \{deebul.nair, paul.ploeger\}@h-brs.de}
}

\keywords{ Automatic Short Answer Grading, Transfer Learning, ELMo, BERT, GPT, GPT-2}
%The paper must have at least one keyword. The text must be set to 9-point font size and without the use of bold or italic font style. For more than one keyword, please use a comma as a separator. Keywords must be titlecased.

\abstract{\ac{ASAG} is the process of grading the student answers by computational approaches given a question and the desired answer. Previous works implemented the methods of concept mapping, facet mapping, and some used the conventional word embeddings for extracting semantic features. They extracted multiple features manually to train on the corresponding datasets. We use pretrained embeddings of the transfer learning models, ELMo, BERT, GPT, and GPT-2 to assess their efficiency on this task. We train with a single feature, cosine similarity, extracted from the embeddings of these models. We compare the RMSE scores and correlation measurements of the four models with previous works on Mohler dataset. Our work demonstrates that ELMo outperformed the other three models. We also, briefly describe the four transfer learning models and conclude with the possible causes of poor results of transfer learning models.}

\onecolumn \maketitle \normalsize \setcounter{footnote}{0} \vfill

\section{\uppercase{Introduction}}
\label{sec:introduction}

\begin{figure}[!t]
		\centering
		\includegraphics[width=\columnwidth]{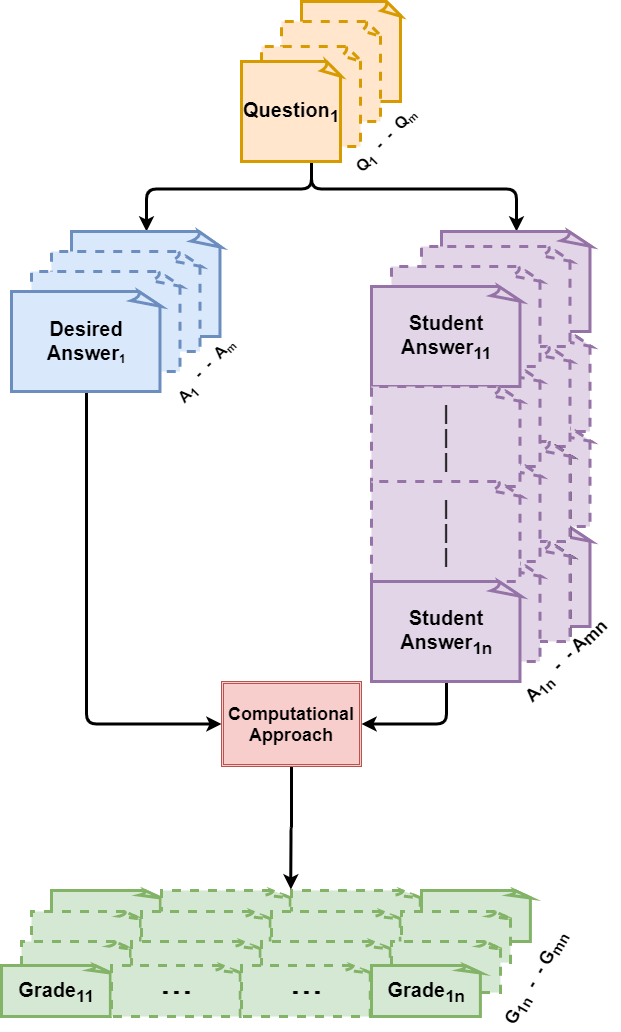}
		\caption{Automatic short answer grading pipeline.}
		\label{fig:ASAG}
	\end{figure}

\noindent 
	 Descriptive examinations test students' comprehensive understanding of topics. With the growing number of students on online platforms and in universities, it is very strenuous to evaluate all the answers manually. The process of grading these detailed answers automatically without any human intervention is achieved by \ac{ASAG}. \ac{ASAG} evaluates the student's answer, given a question and the desired answer as shown in Figure \ref{fig:ASAG}.
	 	
	\cite{perez2005effects}, \cite{bukai2006automated}, implemented the corpus-based techniques to evaluate the student answers. \cite{gutl2007examiner}, \cite{bailey2008diagnosing}, \cite{hou2011automatic} introduced machine learning techniques into \ac{ASAG}. \cite{mohler2011learning} had provided methods extracting syntactic, lexical, morphological, and semantic features for the task. Previous works have manually combined some of the aforementioned language features to train on a regression model. One of such features is the semantic similarities between the desired answer and the student answer. This can be considered as the \ac{STS} task, which assigns the similarity score between two textual corpora. 
	
	The latest advancements in \ac{NLP} and deep learning have provided propitious methods and architectures \cite{bahdanau2014neural} \cite{vaswani2017attention} resulting in robust transfer learning models, to solve multiple tasks. The transfer learning  models were pretrained on huge corpora and are able to extract the semantic context of the words with robust architectures as we explain briefly further in Sec \ref{sec:transfer_learning_models}. These models include \ac{ELMo} \cite{peters2018deep}, \ac{BERT} \cite{devlin2018bert}, \ac{GPT} \cite{radford2018improving} and GPT-2 \cite{radford2019language},  which had shown the state-of-the-art results on various tasks.
	
	We use the embeddings of these transfer learning models to extract the semantic knowledge from the answers by encoding with contextual vectors. We also use fewer preprocessing steps to train a regression model compared to previous works, to assess the potentiality of transfer learning models. We evaluate the efficiency of the embeddings of these models on the Mohler dataset \cite{mohler2011learning}. This provides the effectiveness of using embeddings of pretrained transfer learning models on the task of \ac{ASAG}.
	
	Further, we discuss the related work in Sec. \ref{sec:related_work}. Sec. \ref{sec:transfer_learning_models} briefly explains the transfer learning models and Sec. \ref{sec:dataset} elaborates the dataset. Sec. \ref{sec:experimentation} details the experimentation followed by the results in Sec.\ref{sec:results}. We discuss our observations from the results on \ac{ASAG} in Sec. \ref{sec:observations} and conclude in the Sec. \ref{sec:conclusion}. Finally, Sec. \ref{sec:future_work} explains the future work.

\section{\uppercase{Related work}}
\label{sec:related_work}

	\cite{burstein1999using}, \cite{callear2001caa}, \cite{wang2008assessing} worked with concept mapping techniques, by mapping the related concepts of student answer with desired answer. Later the works of \cite{mitchell2002towards}, \cite{bachman2002reliable} \cite{thomas2003evaluation}, extracted the information from the answers through pattern matching. They have used regular expressions and parse trees for extracting the patterns. \cite{perez2005effects} introduced the corpus-based methods to \ac{ASAG} with the combination of \ac{LSA} and \ac{BLEU} scores. \cite{mohler2009text} used the combination of multiple knowledge based and corpus-based features for extracting the similarity between the students' and teacher's answer. This work is followed by the combining corpus-based methods into machine learning systems \cite{mohler2011learning}. 
	
	\cite{sultan2016fast} extracted multiple features such as word-alignment, vector similarity, \ac{tf-idf}, trained and evaluated on the Mohler dataset \cite{mohler2011learning}. \cite{tim2019computer} used the conventional embeddings in \ac{NLP} such as Word2Vec \cite{mikolov2013distributed} \cite{mikolov2013efficient}, GloVe \cite{pennington2014glove} and FastText \cite{bojanowski2017enriching} to extract the distributional and semantic features through embeddings. \cite{tim2019computer} compared the results with pretrained word embeddings and domain-specific trained word embeddings of Word2Vec, GloVe and FastText on Mohler dataset. 
	
	However, the conventional embeddings did not consider the context of the words and long-term dependencies. The later advancements in \ac{NLP} contemplated the contexts of the words in the sentences, pretrained on huge corpora. This resulted in the models like \ac{ELMo} \cite{peters2018deep}, \ac{GPT} \cite{radford2018improving}, \ac{BERT} \cite{devlin2018bert} and GPT-2 \cite{radford2019language}. In addition, the former works tend to use multiple features to evaluate the answers. We use only a single semantic feature, extracted from the transfer learning models to train. 
	
\section{\uppercase{Transfer learning models}}
\label{sec:transfer_learning_models}
	
	\begin{figure*}[!t]
		\centering
		\begin{subfigure}[t]{0.3\textwidth}
			\centering
			\includegraphics[width=\textwidth]{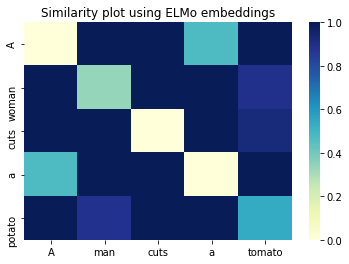}
			\caption{First layer}
			\label{fig:elmo_first_layer}
		\end{subfigure}
		\begin{subfigure}[t]{0.3\textwidth}
			\centering
			\includegraphics[width=\textwidth]{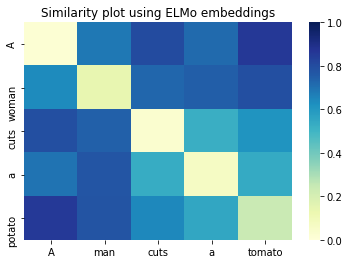}
			\caption{Second layer}
			\label{fig:elmo_second_layer}
		\end{subfigure}
		\begin{subfigure}[t]{0.3\textwidth}
			\centering
			\includegraphics[width=\textwidth]{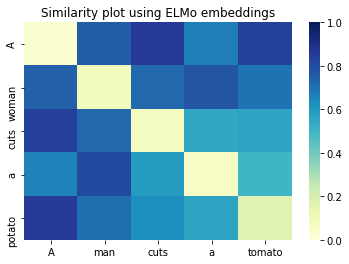}
			\caption{Third layer}
			\label{fig:elmo_third_layer}
		\end{subfigure}
		
		\caption[Transition of semantic information in ELMo]{Transition of semantic information retrieval in stacked LSTM of ELMo. The lighter the color the similar the words are.}
		\label{fig:elmo_semantics}
	\end{figure*}

	We use the embeddings of the pre-trained transfer learning models to extract the semantics of the words based on their context. These transfer learning models are initially trained on a source task and can be used on various target tasks by finetuning the ultimate layers. However, in this work, we do not consider the idea of transfer learning, instead we only use the pretrained embeddings of these models trained on the source task, assuming that they have extracted the significant features of each word from the pretrained huge corpora. The brief explanation of the transfer learning models in creating their pretrained embeddings are explained in the subsequent paragraphs.
	
	\paragraph{\ac{ELMo}} was built with three layers of \ac{LSTM}, similar to that used by \cite{jozefowicz2016exploring} \cite{kim2016character} for complex language modeling. \ac{ELMo} extracts the bi-directional context of the words by concatenating the joint probabilities of forward language model and backward language model. This models assists highly for homonyms\footnote{Homonyms are the words having same spelling and pronunciation but different meanings} such as \textit{play}, \textit{train} and \textit{spring}. ELMo assigns different words embeddings for the same word in different contexts. The final layers of stacked LSTM are good at retrieving semantics, while	the initial layers are better at extracting syntactic information. The transition of syntactic to semantic information retrieval can be visualized in Figure \ref{fig:elmo_semantics}.	
	
	\paragraph{\ac{GPT}} used a stacked transformer architecture \cite{vaswani2017attention} to train the weights. Stacked transformer architecture provides more structured memory for the long-term dependencies compared to recurrent neural networks \cite{radford2018improving}. \ac{GPT} aims to create an effective procedure for transfer learning through semi-supervised approach, with a combination of unsupervised pretraining and supervised finetuning. Principally, a language model objective is applied as a source task on unlabeled data, to learn initial parameters of the neural network. Later, the architecture is finetuned for a required task using corresponding supervised traversal-style approaches \cite{rocktaschel2015reasoning}. The traversal-style approach creates a single contiguous sequence of tokens for a structured text.
		
	\paragraph{\ac{BERT}} extracts the benefits of both ELMo and GPT, resulting
	in using the transformer mechanism \cite{vaswani2017attention} and capturing bi-directional context. \ac{BERT} is trained on a multi-layer bidirectional transformer encoder with a huge corpus of BookCorpus \cite{zhu2015aligning} and Wikipedia datasets. The authors have not adopted the traditional language modeling as source task, because it is only possible to train language models either left-to-right or right-to-left, which seemed to lose the essence of capturing the bi-directional context effectively \cite{devlin2018bert}. Instead, they introduced an approach called \ac{MLM} as a source task. \ac{MLM} masks some tokens in the sentences and predicts the tokens during training. Bert was also pretrained on 'Next Sentence Prediction' task.
	
	\paragraph{GPT-2} aims to show that language models can learn multi-tasks through unsupervised learning. The former GPT model used a combination of unsupervised pretraining and supervised finetuning. Language modeling is the core part of GPT-2. The authors used the transformer architecture with the similar architecture of GPT with minimal changes. They extracted the WebText dataset from web scraping, considering it to be more generalized. An overview of the four transfer learning models' architectures and datasets are represented in the Table \ref{tab:transfer_learning_models}.
	
	\begin{table*}[!h]
		\centering		
		\caption{An overview of pretrained transfer learning models}
		\begin{tabular}{|l|l|l|l|}
			\hline			
			\textbf{Model} & \textbf{Architecture} & \textbf{Pretrained dataset} & \textbf{Dataset size} \\	\hline
			
			ELMo & Stacked bi-LSTM & One billion word benchmark & 1B words \\ \hline
			
			GPT & Stacked transformer & BookCorpus & 800M words \\ \hline
			
			BERT & Stacked transformer & \begin{tabular}[c]{@{}l@{}}BookCorpus \\ English Wikipedia\end{tabular} & \begin{tabular}[c]{@{}l@{}}800M words\\ 2500M words\end{tabular} \\ \hline
			
			GPT2 & Stacked transformer & WebText & \textit{Not known} \\ \hline
		\end{tabular}

		\label{tab:transfer_learning_models}		
	\end{table*}

\section{\uppercase{Dataset}}
\label{sec:dataset}

	Mohler dataset comprises of questions and answers in computer science domain \cite{mohler2011learning}. The goal of the dataset is to evaluate the model in grading the students' answers by comparing them with the evaluator's desired answer. It constitutes 2273 answers from 10 assignments and 2 examinations, collected from 31 students for 80 different questions. 
	
	Each answer in the assignment is graded from 0 (not correct) to 5 (totally correct) by two evaluators, who are specialized in the field of computer science. The average of the two evaluators' scores is considered as the standard score of each answer. The answers in examinations are graded from 0 (not correct) to 10 (totally correct). To eliminate the clutter of assigned scores in the dataset, \cite{tim2019computer} had normalized all the grades of examinations to 0-5. Hence, we intend to use this cleaned data in our experimentation and evaluation.

\begin{figure}[!t]
		\centering
		\includegraphics[width=\columnwidth]{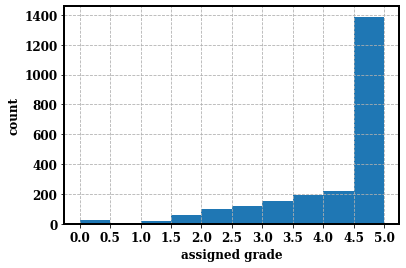}
		\caption{Histogram of assigned average scores in Mohler dataset representing the bias towards correct answers.}
		\label{fig:mohler_score_histogram}
	\end{figure}

	The dataset is biased towards correct answers \cite{mohler2011learning} with the mean of average grades being 4.17 and median being 4.50. This bias can be inferred from Figure \ref{fig:mohler_score_histogram}. 
	
\section{\uppercase{Experimentation}}
\label{sec:experimentation}

\paragraph{Preprocessing} We have conducted only a tokenization in the preprocessing step. Lemmatization and stop word removal are neglected consciously, to analyze the performance of transfer learning models in raw setup. \cite{tim2019computer} also used the spell checker for correcting the misspelled words, which we have avoided. We have assumed that the graders have deducted grades for the misspelled words in the students' answers. Henceforth, the existent spelling mistakes may be trained as a negative feature\footnote{Here by negative feature, we mean the learnt feature that results in adverse effects.}, internally. Since these transfer learning models are trained on huge vocabulary, it is plausible to assume that they can understand the misspelled words to an extent. The versatility of transfer learning models to assign an embedding to the new words also assisted in disregarding the spell mistakes.

\paragraph{Feature extraction} The pretrained embeddings of each transfer learning model are assigned to the tokens of each word in all the answers. Considering there exists $n_j$ words for an answer $j$ of question $i$, we create answer embeddings by \ac{SOWE} as depicted in Eq \ref{eq:sowe}, where $a_{ij}$ represents the $j^{th}$ answer vector of question $q_i$, $w_{k}$ represents the vector of the $k^{th}$ word in the answer $a_{ij}$. This creates a single vector representing each answer in high dimensional hypothesis space. The size of the sentence embeddings is equal to the size of the word embeddings.

\begin{equation}
\label{eq:sowe}
	a_{ij} = \sum_{j=1}^{n_j}w_{k}
\end{equation}

We calculate the similarity between each student answer $a_{ij}$ and desired answer $a_i$ with cosine similarity given in Eq \ref{eq:cosine_similarity}. We normalize these scores from 0 to 1 to scale the similarities and attain the relative measure of similarities. We consider these scores as the features of the answers and train them with different regression methods.

\begin{equation}
\label{eq:cosine_similarity}
cos(a_{ij},a_i) = \frac{a_{ij}.a_i}{|a_{ij}||a_i|}
\end{equation}

\paragraph{Training and Testing} We randomly split the Mohler data to 70\% of training and 30\% of testing data. We train each model for 1000 iterations selecting different training and testing data randomly for every iteration to generalize the results. We train the cosine similarity feature with the correspondingly assigned grades with isotonic, linear and non-linear (ridge) regression. We implement the selected regression models to compare our results with \cite{mohler2011learning} and \cite{tim2019computer}. Followed by training, we test the trained regression model on test data. This test data is not seen by the regression model until it's testing phase. The test data's similarity scores are input through the trained regression model. This results in the predicted grades, which will be further used for evaluation. During evaluation, we calculate the \ac{RMSE} and Pearson correlation between the predicted scores and desired scores.

\section{\uppercase{Results}}
\label{sec:results}

	\begin{table*}[!t]
		\centering
		\caption{Root Mean Square Error (RMSE) and Pearson correlation ($\rho$)  of pretrained transfer learning models on Mohler dataset}
		\begin{tabular}{|c|cc|c|c|c|c|}
			\hline
			\multirow{2}{*}{\textbf{Model}} & \multicolumn{2}{c|}{\textbf{Isotonic regression}}    & \multicolumn{2}{c|}{\textbf{Linear regression}} & \multicolumn{2}{c|}{\textbf{Ridge regression}} \\ \cline{2-7} 
			& \multicolumn{1}{c|}{\textbf{RMSE}} & \textbf{$\rho$} & \textbf{RMSE}          & \textbf{$\rho$}        & \textbf{RMSE}         & \textbf{$\rho$}        \\ \hline
			\textbf{ELMo}                   & \textbf{0.978}                     & \textbf{0.485}  & \textbf{0.995}         & \textbf{0.451}         & \textbf{0.996}        & \textbf{0.449}         \\
			\textbf{GPT}                    & 1.082                              & 0.248           & 1.088                  & 0.222                  & 1.089                 & 0.217                  \\
			\textbf{BERT}                   & 1.057                              & 0.318           & 1.077                  & 0.266                  & 1.075                 & 0.269                  \\
			\textbf{GPT-2}                  & 1.065                              & 0.311           & 1.078                  & 0.274                  & 1.079                 & 0.269                  \\ \hline
		\end{tabular}
		\label{table:results}
	\end{table*}
	
	\begin{table*}[!h]
		\centering
		\caption[Overview comparison of results]{Overview comparison of results on Mohler dataset with former approaches}
		\begin{tabular}{|c|l|c|c|}
			\hline
			\textbf{Model/Approach} & \multicolumn{1}{c|}{\textbf{Features}} & \textbf{RMSE} & \textbf{Pearson correlation} \\ \hline
			\multirow{2}{*}{\textbf{BOW \cite{mohler2011learning}}} & SVMRank & 1.042 & 0.480 \\ \cline{2-4} & SVR & 0.999 & 0.431 \\ \hline
			{\textbf{tf-idf \cite{mohler2011learning})}} & SVR & 1.022 & 0.327 \\ \hline
			\multicolumn{1}{|l|}{\textbf{tf-idf \cite{sultan2016fast}}} & LR + SIM                               & \textbf{0.887} & \textbf{0.592}               \\ \hline
			\multirow{2}{*}{\textbf{Word2Vec \cite{tim2019computer}}}       & SOWE + Verb phrases                    & 1.025          & 0.458                        \\ \cline{2-4} 
			& SIM+Verb phrases                       & 1.016          & 0.488                        \\ \hline
			\multirow{2}{*}{\textbf{GloVe \cite{tim2019computer}}}          & SOWE + Verb phrases                    & 1.036          & 0.425                        \\ \cline{2-4} 
			& SIM+Verb phrases                       & 1.002          & 0.509                        \\ \hline
			\multirow{2}{*}{\textbf{FastText \cite{tim2019computer}}}       & SOWE + Verb phrases                    & 1.023          & 0.465                        \\ \cline{2-4} 
			& SIM+Verb phrases                       & 0.956          & 0.537                        \\ \hline
			\textbf{ELMo}                                 & SIM                                    & 0.978          & 0.485                        \\ \hline
			\textbf{GPT}                                  & SIM                                    & 1.082          & 0.248                        \\ \hline
			\textbf{BERT}                                 & SIM                                    & 1.057          & 0.318                        \\ \hline
			\textbf{GPT-2}                                & SIM                                    & 1.065          & 0.311                        \\ \hline
		\end{tabular}
		\label{table:overview_results}
	\end{table*}

	Table \ref{table:results} depicts the RMSE and Pearson correlation ($\rho$) results of pretrained embeddings of transfer learning models on Mohler dataset. The \ac{RMSE} score defines the absolute error between the desired and predicted scores. Henceforth, the lower the \ac{RMSE}, the better the model is. Pearson correlation measures the comprehensive correspondence between the assignment of desired scores and predicted scores. Therefore, the higher the $\rho$, the better the model is. Table \ref{table:overview_results} provides a comparison of our results with former approaches and models. 		
	
	\ac{ELMo} embeddings attained a Pearson correlation of 0.485 and a \ac{RMSE} score of 0.978 with isotonic regression. The results on linear and ridge regression are 0.451 and 0.449 for Pearson correlation; 0.995 and 0.996 for \ac{RMSE} score respectively. \ac{GPT} embeddings resulted in Pearson correlation of 0.248, 0.222 and 0.217\footnote{	We always provide our results in the order of isotonic, linear and ridge regressions when three consecutive results are detailed.}. The results of \ac{RMSE} on \ac{GPT} embeddings are 1.082, 1.088 and 1.089.
	
	\ac{BERT} embeddings also performed better with isotonic regression with a $\rho$ of 0.318 and \ac{RMSE} of 1.057 compared to linear ($\rho$ : 0.266; \ac{RMSE}: 1.077) and non-linear ($\rho$ : 0.267; \ac{RMSE}: 1.075) regressions. GPT-2 has performed alike \ac{BERT} with $\rho$ values of 0.311, 0.274 and 0.269 and RMSE scores of 1.065, 1.078 and 1.079 respectively on three regression training models.
	
	Table \ref{table:overview_results} compares our results with conventional embeddings of Word2Vec, GloVe and FastText. Also, the traditional approaches of \ac{BOW} and tf-idf on Mohler dataset are compared. However, we only consider the models or approaches that do not use domain-specific training of the data, for legible comparison. Various features or algorithms used by the models are demonstrated in the \textit{Features}\footnote{SVM- Support Vector Machine; SVR - Support Vector Regression; LR- Length Ratio between desired answer and student answer; SIM - Similarity score; SOWE - Sum of Word Embeddings} column of the Table \ref{table:overview_results}. The Pearson correlation of 0.592 and \ac{RMSE} of 0.887 illustrate that tf-idf approach considering the combination of length ratio and similarity features by \cite{sultan2016fast} outperformed other approaches.

\section{\uppercase{Observations}}
\label{sec:observations}

	Our results illustrate that ELMo model outperformed on domain-specific \ac{ASAG} compared to other transfer learning models. The reasons that ELMo may have worked better than the other transfer learning models on the domain-specific dataset is two-fold. Firstly, the assignment of different vectors for the same word in different contexts. This is helpful for the homonyms. Secondly, the availability of significant amount of domain data in the pre-trained corpus compared to the other transfer learning models. The pretrained data of \ac{BERT}, \ac{GPT} and GPT-2 models is extensive and the domain which we are testing is comparatively very smaller. This resulted in the similarity scores in the range of $10^{-5}$ to $10^{-1}$.
	
	Isotonic regression worked better than the linear and ridge regressions. This is because, the isotonic regression trains step-wise, a similar way of assigning grades to the students manually, unlike linear and non-linear regression methods. We also noted that linear and ridge regression results of transfer learning models' were almost similar. This may be due to the significant linear fit of the data and the negligible non-linearity between the grade and the trained feature.
	
	Compared to the other former approaches, considering the several preprocessing steps followed by multiple feature extraction and training, ELMo embeddings demonstrate competing results, without any preprocessing or multiple feature training. 

\section{\uppercase{Conclusion}}
\label{sec:conclusion}
	
	We evaluated the embeddings of four transfer learning models on Mohler dataset (domain-specific) on the task of \ac{ASAG}. These transfer learning models (\ac{ELMo}, \ac{GPT}, \ac{BERT} and GPT-2) were explained breifly with their pretraining procedures. Besides, we also elucidated the \ac{ASAG} task's significance and it's applications. The sentence embeddings are created from all the selected four transfer learning models for all the desired and student answers of the dataset. The encoding of answers are related to the words in the answers, irrespective of their order. The cosine similarity feature was extracted for every student	answer and desired answer. This feature was trained with all the three isotonic, linear and ridge regression methods.
	
	\ac{ELMo} outperformed other transfer learning models on the task with a best \ac{RMSE} score of 0.978 and Pearson correlation of 0.485. With these results, \ac{ELMo} competed with the conventional word embeddings, such as Word2Vec, GloVe and FastText, without any preprocessing or multiple feature training. \ac{ELMo} performed comparatively better than other transfer learning models, \ac{BERT}, \ac{GPT} and GPT-2. These transfer learning models have exhibited poor results on the Mohler dataset compared to the conventional word embeddings. We also concluded that ELMo can achieve near to the state of the art results without any further training of domain-specific data or compelling preprocessing of data.
	
\section{\uppercase{Future Work}}
\label{sec:future_work}

	Although we have achieved significant results with ELMo without any preprocessing or multiple feature extraction or training, there is much scope to extend this work further. Firstly, it is important to consider the question demoting and elimination of stop words as in \cite{mohler2011learning}, \cite{sultan2016fast} and \cite{tim2019computer}. The word alignment procedure by \cite{sultan2016fast}, can have a substantial effect on sentence embeddings as most of the insignificant words can be removed.
	
	Moreover, it is also important to explore different methods of assigning sentence embeddings such as mean-pooling, max-pooling, other than the sum of the vectors. The use of explicit sentence embeddings such as universal sentence encoders\cite{cer2018universal} can be effective to create a similarity feature. From the obtained results, it is also promising to use transfer learning embeddings trained or finetuned on domain-specific data. It is also necessary to update partisan datasets to be unbiased for better experimentation and evaluation.

\section*{\uppercase{Acknowledgements}}

	We thank Tim Metzler for providing the updated version of the Mohler dataset for experimentation. 

\bibliographystyle{apalike}
{\small \bibliography{evaluation_on_mohler}}

\end{document}